\documentclass{article}
\usepackage{ijcai26}

\usepackage{times}
\usepackage{soul}
\usepackage{url}
\usepackage[hidelinks]{hyperref}
\usepackage[utf8]{inputenc}
\usepackage[small]{caption}
\usepackage{graphicx}
\usepackage{amsmath}
\usepackage{amsthm}
\usepackage{booktabs}
\usepackage{algorithm}
\usepackage{algorithmic}
\usepackage[switch]{lineno}

\usepackage{amssymb}
\usepackage{bm}
\usepackage{dsfont}
\usepackage{arydshln}
\usepackage{dblfloatfix}
\usepackage{multirow}
\usepackage{natbib}
\usepackage{CJKutf8}
\usepackage{xcolor}

\title{Towards Fine-Grained Code-Switch Speech Translation \\ with Semantic Space Alignment}

\author{
Yan Gao$^1$\thanks{These authors contributed equally.}\and
Yazheng Yang$^1$\footnotemark[1]\and
Zhibin Lan$^1$\and
Yidong Chen$^1$\and
Min Zhang$^2$\and
Daimeng Wei$^2$\and \\
Derek F. Wong$^3$\And
Jinsong Su$^1$\thanks{Corresponding author.}
\\
\affiliations
$^1$School of Informatics,  Xiamen University, China $^2$Huawei Translation Services Center, Beijing, China\\
$^3$NLP$^2$CT Lab, Department of Computer and Information Science, University of Macau\\
\emails
gaoyan@stu.xmu.edu.cn
}

\begin{document}

\maketitle

\begin{abstract}
Code-switching (CS) speech translation (ST) aims to translate speech that alternates between multiple languages into a target language text, posing significant challenges due to the complexity of semantic modeling and the scarcity of CS data. Previous studies mainly rely on the models themselves to implicitly learn semantic representations and resort to costly manual annotations. To mitigate these limitations, we propose enhancing Large Language Models (LLMs) with a Mixture-of-Experts (MoE) speech projector composed of language expert groups, where each group specializes in the semantic space of a specific language for fine-grained speech feature modeling. A language-specific loss and an intra-group load balancing loss are jointly introduced to guide efficient token routing across and within expert groups. Furthermore, we introduce a multi-stage training paradigm that utilizes readily available automatic speech recognition (ASR) and monolingual ST data, facilitating speech-text alignment and improving translation performance. To bridge the data gap for smooth domain transfer, a transition loss is employed to improve adaptation to CS scenarios. Extensive experiments on widely used datasets demonstrate the effectiveness and generality of our approach, achieving average improvements of $0.86$ BLEU and $0.93$ COMET over SeamlessM4T, with maximum improvements of $1.49$ BLEU and $1.41$ COMET across different test sets.\footnote{Our code is available at \url{https://github.com/XMUDeepLIT/CSST-SSA}.}
\end{abstract}

\section{Introduction}
Code-switching (CS), the practice of alternating between multiple languages within a single utterance or discourse, is a pervasive phenomenon in multilingual speech and poses significant challenges to speech and language processing models~\citep{scotton1977bilingual}. The task of CS speech translation (ST) aims to translate such speech into text in a target language. With globalization~\citep{winata-etal-2023-decades}, CS has become increasingly prevalent, extending beyond multilingual communities to predominantly monolingual settings, thus drawing much attention recently.

While CS has been extensively explored in machine translation (MT)~\citep{xu-yvon-2021-traducir,gupta-etal-2021-training,vavre-etal-2022-adapting,gaser-etal-2023-exploring,pengpun2024creating,borisov-etal-2025-low} and automatic speech recognition (ASR)~\citep{winata-etal-2020-meta,chi-bell-2022-improving,dhawan-etal-2023-unified,kronis-etal-2024-code}, its research in ST remains under-explored. This is primarily due to two key challenges: 1) ~$\emph{semantic modeling complexity}$ introduced by language alternation, which poses significant challenges for the model in capturing effective representations from CS speech; 2)~$\emph{data scarcity}$, as high-quality and large-scale CS speech translation datasets are limited and difficult to construct, which limits effective model training. Prior methods tend to ignore the semantic complexity arising from language switching, simply leaving it to the models themselves to implicitly learn and resolve during training~\citep{huber2022code,weller2022end,alastruey2023towards,p-s-v-n-etal-2025-costa}, which hinders overall performance. To mitigate data scarcity, previous studies resort to manual annotations~\citep{alastruey2023towards,p-s-v-n-etal-2025-costa}, which are inefficient and costly, producing datasets of limited scale. Other approaches focus on synthetic data, which often suffer from unnatural switching, disrupted word order, and grammatical inconsistencies~\citep{xie2025switchlingua}.

Recently, large language models (LLMs) have demonstrated remarkable performance across diverse tasks. To equip LLMs with cross-modal capabilities, prior approaches typically introduce a projector that connects a pretrained speech encoder to the LLM~\citep{chen-etal-2024-llast, zhang2024multi, wu2025locate, li2026plast}. This architecture effectively integrates the speech encoder’s strength in extracting acoustic features with the LLM’s powerful language modeling ability. However, existing studies reveal that despite being pretrained on large-scale multilingual corpora, LLMs still exhibit limited cross-lingual capability, resulting in inferior performance on CS data relative to monolingual data~\citep{zhang2023multilingual,mohamed2025lost}. Moreover, current research on LLMs in the context of CS primarily focuses on MT~\citep{zhang2024code,gupta2024codemixeryanahinovel}, leaving their potential in CS speech translation largely unexplored.

In this work, to address the aforementioned challenges, we explore the use of LLMs for CS speech translation, improving performance from the perspectives of both model architecture and data utilization. 1) Based on the above-mentioned architecture, we introduce a novel Mixture-of-Experts (MoE)~\citep{shazeer2017outrageously} speech projector, where experts are organized into language-aware groups to enable fine-grained modeling of speech features from different languages, thereby better capturing the complex cross-lingual semantics in CS speech. 2) To mitigate the scarcity of high-quality CS speech translation data, we propose a novel multi-stage training paradigm that leverages high-resource ASR data to strengthen semantic comprehension and monolingual ST data to establish robust translation capability, followed by a gradual transition to CS speech translation data that enables smooth domain adaptation. Specifically, our training paradigm comprises four stages. In the first stage, ASR data are used to pretrain individual projectors for each language involved in the target CS scenario. In the second stage, these projectors of different languages form individual expert groups, which are further aggregated to construct a MoE speech projector. Additionally, a language-specific loss and an intra-group load balancing loss are introduced to better guide the learning of the MoE speech projector and promote effective expert specialization. The first two stages jointly align the speech and text modalities, facilitating the learning of cross-modal representations. In the third stage, we progressively transition from ASR to monolingual ST data using a transition loss to enhance the model’s translation ability. In the final stage, the model is further adapted from monolingual ST to CS speech translation data, thus enabling effective translation of CS speech.

To summarize, the main contributions of our work include the following four contributions:
\begin{itemize}
\setlength{\itemsep}{2pt}
\setlength{\parsep}{2.5pt}
\setlength{\parskip}{0pt}
\item{To the best of our knowledge, we first explore LLMs in end-to-end CS speech translation, providing baseline results to facilitate future research.}

\item{We propose a novel MoE speech projector that enables fine-grained modeling of speech features from different languages, thereby enhancing the model’s ability to capture semantic distinctions in CS speech.}

\item{We propose a multi-stage training paradigm that progressively guides the model to learn speech-text alignment and adapt to CS scenarios, effectively mitigating the scarcity of high-quality CS speech translation data.}

\item{Empirical evaluations on monolingual and CS speech translation datasets validate the effectiveness of our model, and in-depth analysis experiments provide detailed insights into the contribution of each module.}
\end{itemize}

\section{Related Works}
CS translation in natural language processing (NLP) has garnered much attention and focuses mainly on MT~\citep{xu-yvon-2021-traducir,gupta-etal-2021-training,vavre-etal-2022-adapting,gaser-etal-2023-exploring,pengpun2024creating,borisov-etal-2025-low}. However, CS frequently occurs in spoken contexts such as lectures, meetings, and phone conversations, which necessitates handling speech input. To handle speech input, such methods are typically cascaded with ASR models~\citep{winata-etal-2020-meta,chi-bell-2022-improving,dhawan-etal-2023-unified,kronis-etal-2024-code}, which suffer from error propagation. Recent research efforts explore end-to-end (E2E) ST to mitigate this issue~\citep{weller2022end,huber2022code,alastruey2023towards,yang2024investigating,p-s-v-n-etal-2025-costa}. A representative approach, COSTA~\citep{p-s-v-n-etal-2025-costa}, enhances translation quality by aligning speech and text representations via interleaving their respective embeddings during fine-tuning.

Recently, LLMs have demonstrated impressive performance across diverse tasks, leading to increasing research on their use in CS machine translation~\citep{zhang2023multilingual,zhang2024code,gupta2024codemixeryanahinovel,mohamed2025lost} with promising results. Despite these successes, the potential of LLMs for CS speech translation remains largely unexplored. 
Current studies on monolingual ST generally integrate a pretrained speech encoder with an LLM through a projector~\citep{chen-etal-2024-llast, wu2025locate, li2026plast}, thereby equipping the LLM with cross-modal understanding abilities. This framework effectively combines the strength of the speech encoder in extracting speech features with the powerful translation capability of the LLM. Although this framework has shown promising results in monolingual settings, its effectiveness in more challenging CS scenarios remains unclear.

\begin{figure}[t]
	\centering
	\includegraphics[width=0.83\columnwidth]{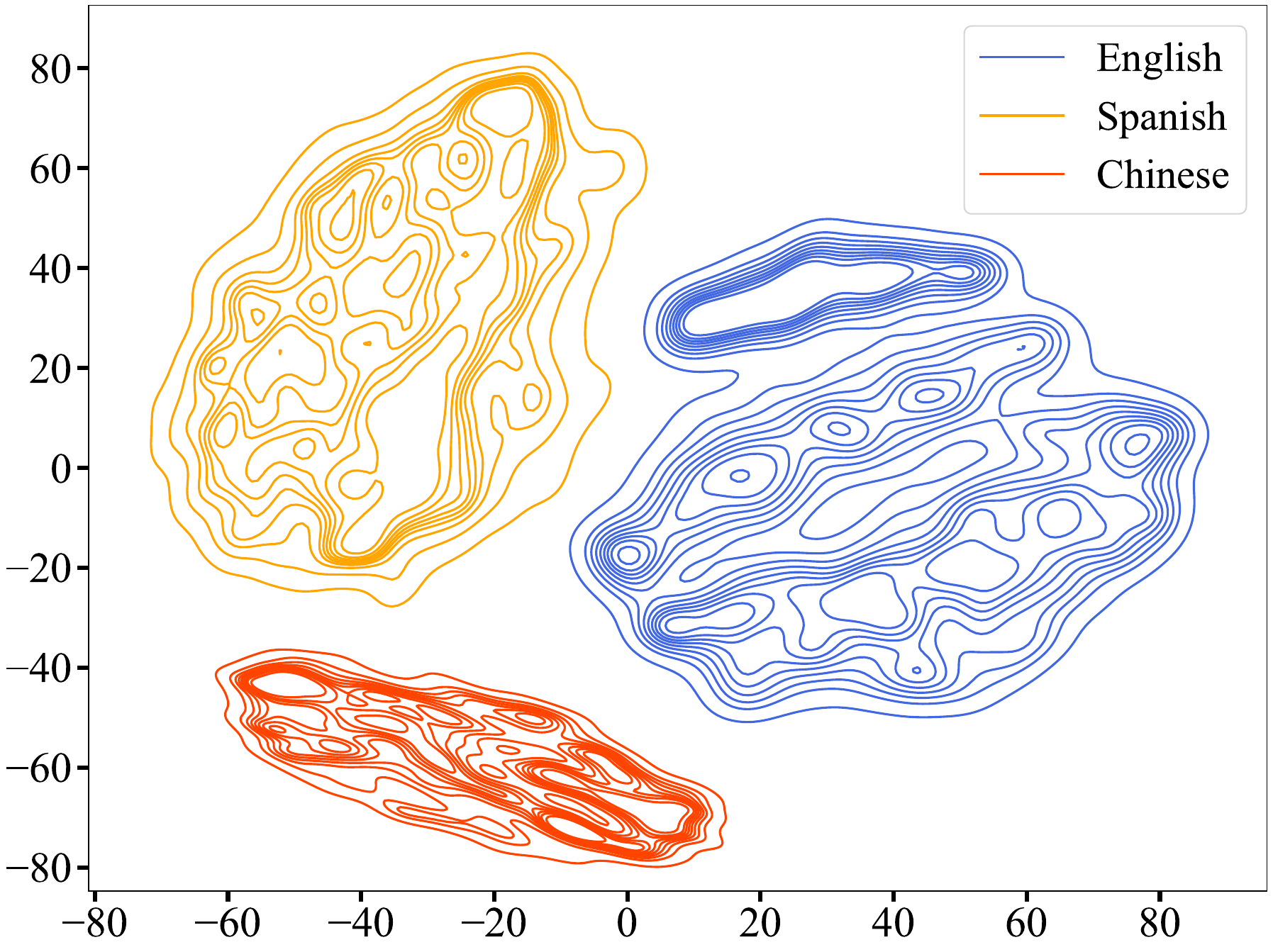}
 	\caption{KDE plots of speech features from the validation set for different languages using t-SNE. }\label{tSNE_KDE}
\end{figure}

\begin{table}[!t]
	\renewcommand
	\arraystretch{1.12}
	\centering
\small
\begin{tabular}{l|ccc}
\hline
\textbf{Settings} & \textbf{En} & \textbf{Es} & \textbf{zh-CN}\\ 
\hline
Share Projector & $7.07$ & $6.01$ & $14.26$ \\
Expert Projector& $7.05$ & $4.52$ & $13.21$ \\
\hline
    \end{tabular}
\caption{
    WER$\downarrow$(En, Es) and CER$\downarrow$(zh-CN) scores of the two settings on the validation set. 
	}\label{tab_pre}
\end{table}

\section{Preliminary Study}
In this section, we explore the complexity of semantic modeling in CS speech from two perspectives. First, we examine whether a semantic space gap exists between different languages. Second, we evaluate the impact of employing a shared projector on model performance.

\begin{figure*}[t]
	\centering
	\includegraphics[width=\textwidth]{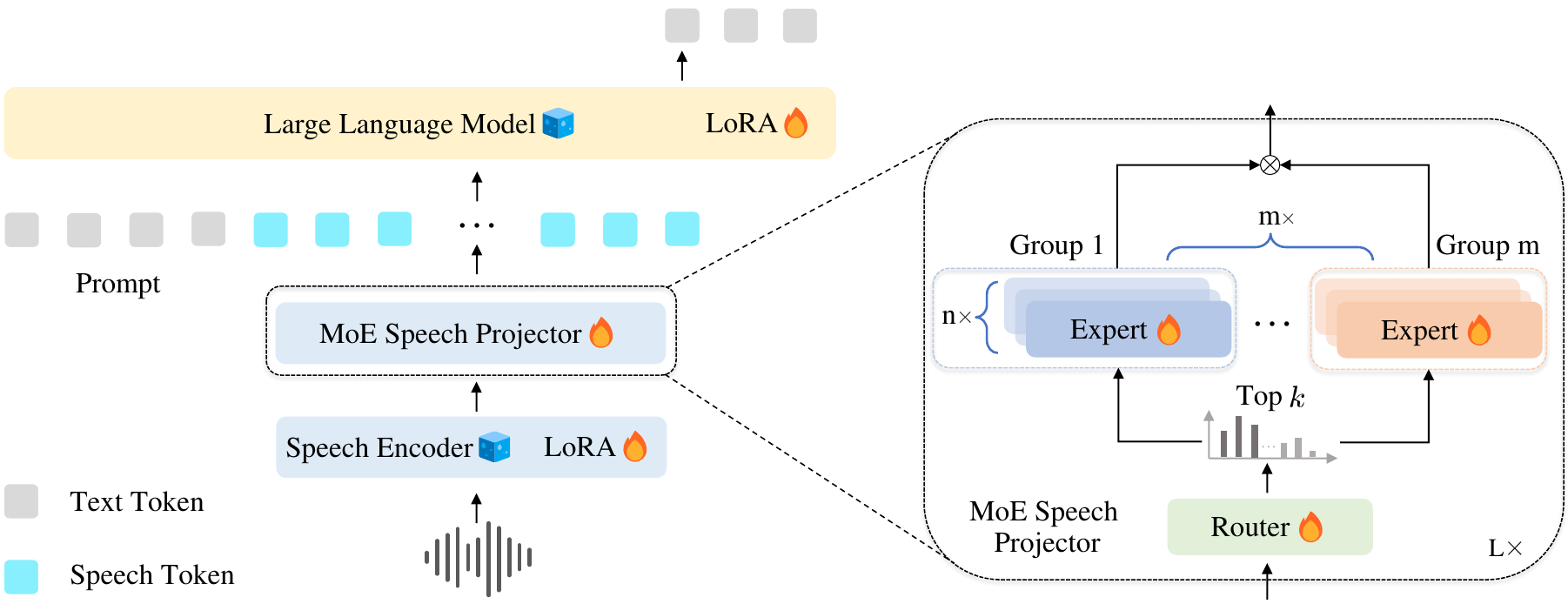}
 	\caption{Illustration of the proposed framework. For CS speech involving $m$ different languages, the MoE speech projector assigns one expert group comprising $n$ experts per language, thus the total number of experts $N=n \times m$. During training, LoRA~\citep{hu2022lora} is applied to fine-tune the parameters of the LLM and the speech encoder, while fully fine-tuning all parameters of the MoE speech projector. }\label{framework}
\end{figure*}
\paragraph{Does a semantic space gap exist between different languages?}
To investigate this issue, we conduct a preliminary experiment using Spanish, English, and Chinese ASR data from the Common Voice corpus~\citep{ardila-etal-2020-common}. Specifically, mean-pooled speech features extracted from the Whisper encoder~\citep{pmlr-v202-radford23a} are reduced to two dimensions using t-SNE~\citep{hinton2002stochastic} and visualized via a bivariate kernel density estimation (KDE) plot, as illustrated in Figure~\ref{tSNE_KDE}. The visualization reveals that samples from the same language cluster tightly, while those from different languages are clearly separated, indicating clear boundaries between languages. These findings provide empirical evidence of a substantial semantic space gap across languages in speech features.

\paragraph{Does sharing the projector across languages affect performance?}
Previous studies typically employ a single projector to align the semantic spaces of the speech encoder and the LLM. However, given the observed cross-lingual semantic space gap, the suitability of a shared projector in capturing semantic distinctions across languages remains uncertain. To investigate this issue, we compare two settings: (1) a shared projector for all languages, in which all speech tokens are processed by the single projector; and (2) one expert projector per language, where each speech token is processed by the projector corresponding to its language. We only fine-tune the projector, while keeping the parameters of the speech encoder and the LLM frozen. Since CS speech lacks oracle language labels at the token level, we also evaluate on monolingual ASR data, where all speech tokens are labeled with their corresponding language. First, we pre-train a single projector on the ASR data to mitigate variance from random initialization and establish a basic semantic alignment. Both the shared and expert projectors are then initialized with the pre-trained weights. We subsequently fine-tune both settings on the same ASR data. The experimental results, reported in Table~\ref{tab_pre}, show that the expert projector consistently outperforms the shared projector, especially for Spanish and Chinese. This finding indicates that using expert projectors for different languages better captures semantic distinctions. However, in practical CS scenarios, oracle language labels for speech tokens are unavailable. To address this limitation, we adopt the MoE projector, enabling the model to route speech tokens to appropriate expert projectors in an adaptive manner.

\section{Methodology}

In this section, we first describe the architecture of our model and then detail the four-stage training strategy, which constructs the MoE speech projector with two auxiliary losses and progressively adapts the model from ASR to monolingual ST and finally to CS speech translation via the transition loss.

\subsection{Model Architecture}
As shown in Figure \ref{framework}, our framework primarily consists of a speech encoder, a MoE speech projector, and an LLM. The speech encoder extracts features from the speech input, which are aligned with the representation space of the LLM via the MoE speech projector. These aligned features are concatenated with the prompt embeddings and fed into the LLM to generate the translation.

\paragraph{Speech Encoder.}
The speech encoder takes a speech input $\bm{s}$ and encodes it into a representation sequence $\bm{h}$:

\begin{equation}
    \bm{h} = \text{Encoder}(\bm{s}),
\end{equation}
where $\bm{h}\in \mathbb{R}^{T_s \times D_{\text{enc}}}$ denotes the encoder output, with sequence length $T_s$ and encoder hidden dimension $D_{\text{enc}}$.

\paragraph{MoE Speech Projector.}
The MoE speech projector connects the speech encoder to the LLM by mapping the representation space of the encoder output to the LLM input. Each MoE layer consists of a set of $N$ expert linear layers $E=\{E_1, E_2, \ldots, E_N\}$ and a sparse router $G$. For the $l$-th MoE layer, the sparse router predicts a probability distribution over experts and selects the top-$k$ experts at the token level. The outputs of the selected experts are then weighted by their corresponding probabilities and summed to produce the output $\bm{h}_{l+1}$. The complete computation process is formulated as follows:

\begin{equation}
    p_i(h^t_l) = \frac{{\exp(G_i(h^t_l))}}{\sum_{i'=1}^{N} \exp(G_{i'}(h^t_l))},
\end{equation}

\begin{equation}
    h^t_{l+1} = \sum_{i \in \text{Top-}k(h^t_l)} p_i(h^t_l) \cdot E_i(h^t_l),
\end{equation}
where $i$ indexes the $i$-th expert, $t$ indexes the $t$-th element in $\bm{h}$, $G(h) = \mathbf{X}_G \cdot h$ and $E(h) = \mathbf{X}_E \cdot h$ denote the linear transformations of the router and an expert, respectively. Specifically, $\mathbf{X}_E^{(1)} \in \mathbb{R}^{D_{\text{enc}} \times D_{\text{LLM}}}$ for the first layer and $\mathbf{X}_E^{(l)} \in \mathbb{R}^{D_{\text{LLM}} \times D_{\text{LLM}}}$ for the remaining layers, where $D_{\text{LLM}}$ denotes the hidden dimension of the LLM.

The overall process is summarized as follows:
\begin{equation}
    \bm{h}_s = \text{MoE Speech Projector}(\bm{h}),
\end{equation}
where $\bm{h}_s\in \mathbb{R}^{T_s \times D_{\text{LLM}}}$ denotes the speech projector output.

\paragraph{LLM.} 
The tokenizer and embedding layer process the prompt text to obtain the representation $\bm{h}_t$, which is concatenated with the speech representation $\bm{h}_s$ to form the LLM input $\bm{h}_z$ :

\begin{equation}
    \bm{h}_z = \bm{h}_t \oplus \bm{h}_s,
\end{equation}
where the operator $\oplus$ denotes vector concatenation, $\bm{h}_t\in \mathbb{R}^{T_t \times D_{\text{LLM}}}$ and $\bm{h}_z\in \mathbb{R}^{(T_t+T_s) \times D_{\text{LLM}}}$, with $T_t$ denoting the sequence length of $\bm{h}_t$.

\subsection{Multi-stage Transition Training}

\paragraph{Stage 1.} 
In the initial stage, we aim to align the speech and text modalities to obtain well-initialized parameters.

Given $m$ languages involved in the CS scenario, we first randomly initialize an MLP-based projector for each language and pretrain it using the corresponding ASR data. The training objective is to minimize the cross-entropy loss:
\begin{equation}
\begin{aligned}
\mathcal{L}_{\text{ce}}=-\sum_{t=1}^{T}{\log p(y_t\mid \bm{s}, \bm{y}_{<t}; \theta)},
\end{aligned}
\end{equation}
where $t$ is the decoding timestep, $y_t$ denotes the target text token at timestep $t$, and $\theta$ refers to the model parameters.

\paragraph{Stage 2.} 
For each language, we initialize an expert group consisting of $n$ experts with the weights of the corresponding pretrained projector to facilitate effective training. The expert groups from all $m$ languages are then aggregated to form the MoE speech projector, with a randomly initialized router inserted between each layer. Thus, the total number of experts is $N = n\times m$. Although all experts within the same group share identical initial weights, the randomly initialized router assigns tokens to different experts, allowing them to gradually specialize through training.

In addition to the cross-entropy loss, we incorporate two auxiliary losses: a language-specific loss and an intra-group load balancing loss. The language-specific loss provides explicit supervision in routing speech tokens to the expert group corresponding to their language. For monolingual ASR data, all speech tokens are labeled with their respective language. Specifically, the loss penalizes the routing of tokens to experts from other language groups and is formally defined as below:

\begin{equation}
\begin{aligned}
\mathcal{L}_{\text{lang}}= -\sum_{t=1}^{T_s} \sum_{l=1}^L \sum_{i=1}^{N} \left[(1 - z_i) \log(1 - p_i(h_l^t)) \right],
\end{aligned}
\end{equation} 
where $L$ is the number of layers, and $z_i$$=$$1$ if expert $i$ belongs to the expert group corresponding to the language of the input speech, otherwise $z_i$$=$$0$.

While the language-specific loss provides supervision at the group level, it lacks regulation over the token distribution among individual experts within each expert group. To address this limitation, the intra-group load balancing loss is introduced to encourage balanced token assignment within each expert group, preventing over-reliance on a subset of experts. The loss is formulated as:
\begin{equation}
\begin{aligned}
\mathcal{L}_{\text{balance}}=\sum_{l=1}^L\sum_{j=1}^m\sum_{i=1}^n{f_{ijl} \cdot P_{ijl}},
\end{aligned}
\end{equation} 
where $f_{ijl}$ is the fraction of tokens in language $j$ dispatched to expert $i$ within the corresponding expert group:
\begin{equation}
\begin{aligned}
f_{ijl} = \frac{\sum_{t\in T_j} \mathds{1}\{\arg\max_{r}\ p_r(h_l^t)=i\}}{\sum_{i'=1}^n\sum_{t\in T_j} \mathds{1}\{\arg\max_{r}\ p_r(h_l^t)={i'}\}},
\end{aligned}
\end{equation}

and $P_{ijl}$ is the average routing probability of assigning tokens in language $j$ to expert $i$ within the corresponding expert group:
\begin{equation}
\begin{aligned}
P_{ijl} = \frac{\sum_{t\in T_j} p_i(h_l^t)}{\sum_{i'=1}^n\sum_{t\in T_j} p_{i'}(h_l^t)},
\end{aligned}
\end{equation} 
where $T_j$ is the set of tokens belonging to language $j$.

Using the same monolingual ASR data as in Stage 1, the overall objective is summarized as follows:
\begin{equation}
\begin{aligned}
\mathcal{L} = \mathcal{L}_{\text{ce}} + \mathcal{L}_{\text{lang}} + \mathcal{L}_{\text{balance}}
\end{aligned}
\end{equation}

\begin{table*}[]
    \setlength{\tabcolsep}{5.5pt}
	\renewcommand
	\arraystretch{1.2}
	\centering
\small
\begin{tabular}{l|rrrrrrrrrrrrrr}
\hline
\multicolumn{1}{c|}{\multirow{2}{*}{\textbf{Dataset}}} & \multicolumn{6}{c}{\textbf{Common Voice}}                                                              & \multicolumn{4}{c}{\textbf{Fisher}}                                  & \multicolumn{4}{c}{\textbf{NTUML2021}}                               \\
\multicolumn{1}{c|}{}                                  & \multicolumn{2}{c}{\textbf{En}} & \multicolumn{2}{c}{\textbf{Es}} & \multicolumn{2}{c}{\textbf{zh-CN}} & \multicolumn{2}{c}{\textbf{CS}} & \multicolumn{2}{c}{\textbf{Mono.}} & \multicolumn{2}{c}{\textbf{CS}} & \multicolumn{2}{c}{\textbf{Mono.}} \\ \hline
Train                                                  & $233.0$K       & $364.4$H       & $64.4$K        & $96.6$H        & $7.1$K          & $10.4$H          & $8.2$K         & $14.7$H        & $130.6$K         & $156.8$H        & $7.9$K          & $5.1$H        & $9.9$K           & $5.7$H          \\
Valid                                                  & $15.5$K       & $26.1$H        & $13.2$K        & $21.8$H        & $4.8$K          & $7.9$H           & $0.2$K         & $0.3$H         & $3.4$K           & $4.1$H          & $1.5$K          & $1.0$H        & $1.5$K           & $0.8$H          \\
Test                                                   & $15.5$K        & $24.7$H        & $13.2$K        & $22.7$H        & $4.9$K          & $8.2$H           & $1.0$K         & $1.6$H         & $10.6$K          & $12.2$H         & $6.2$K          & $4.0$H        & $8.7$K           & $4.9$H          \\ \hline

\end{tabular}
\caption{
    Statistics of various datasets, including the number of pairs in thousands (K) and total duration in hours (H).
	}\label{dataset_statistics}
\end{table*}

\paragraph{Stage 3.} 
This stage focuses on enhancing the translation ability of the model by training on monolingual ST data. However, directly switching from ASR to ST introduces training inconsistency due to the task gap between speech recognition and translation. To enable a smoother transition, we mix ASR data with monolingual ST data and introduce the transition loss:

\begin{equation}
\begin{aligned}
\mathcal{L}_{\text{transition}} = (1-\lambda) \mathcal{L}_{\text{asr\_ce}} + \lambda \mathcal{L}_{\text{st\_ce}},
\end{aligned}
\end{equation} 
where $\lambda=\frac{b}{B}$, $b$ denotes the current batch index, and $B$ is the total number of batches. $\mathcal{L}_{\text{asr\_ce}}$ and $\mathcal{L}_{\text{st\_ce}}$ denote the cross-entropy losses for the ASR and monolingual ST data, respectively.

Additionally, the language-specific loss and the intra-group load balancing loss are also employed in this stage. The overall loss is therefore formulated as follows:

\begin{equation}
\begin{aligned}
\mathcal{L} = \mathcal{L}_{\text{transition}} + \mathcal{L}_{\text{lang}} + \mathcal{L}_{\text{balance}}
\end{aligned}
\end{equation}

\paragraph{Stage 4.} 
In the final stage, we employ CS speech translation data to adapt the translation ability of the model to CS scenarios. To mitigate the distribution gap between monolingual and CS speech translation, we combine monolingual ST with CS speech translation data and apply the transition loss:

\begin{equation}
\begin{aligned}
\mathcal{L}_{\text{transition}} = (1-\lambda) \mathcal{L}_{\text{st\_ce}} + \lambda \mathcal{L}_{\text{csst\_ce}},
\end{aligned}
\end{equation} 
where $\mathcal{L}_{\text{csst\_ce}}$ denotes the cross-entropy loss for the CS speech translation data.

Notably, since the language label of each speech token is unavailable in CS speech, both the language-specific loss and the intra-group load balancing loss are excluded at this stage.

\section{Experiments}
In this section, we conduct extensive experiments to evaluate the effectiveness of our proposed approach. We first describe the experimental setup, including evaluation metrics, datasets, model details, hyperparameters, and baseline methods. We then present the main results and ablation studies to analyze the overall performance and the contribution of each component, followed by in-depth analyses of the experts to better understand their semantic modeling behavior.

\begin{table}[!t]
	\renewcommand
	\arraystretch{1.2}
	\centering
\small
\begin{tabular}{l|cc}
\hline
\textbf{Model} & \textbf{BLEU} & \textbf{COMET}\\ 
\hline
Experts-$3$ & $38.02$ & $80.95$ \\
Experts-$5$ & $38.40$ & $80.62$\\
Experts-$7$ & $\bf{38.55}$ & $\bf{81.02}$\\
Experts-$9$ & $38.35$ & $80.99$\\
Experts-$11$ & $37.74$ & $80.77$\\
\hline
    \end{tabular}
\caption{
    BLEU and COMET scores on the NTUML2021 CS validation set with top-$3$ and varying numbers of experts $n$.
	}\label{expert_performance_n}
\end{table}

\begin{table}[!t]
	\renewcommand
	\arraystretch{1.2}
	\centering
\small
\begin{tabular}{l|cc}
\hline
\textbf{Model} & \textbf{BLEU} & \textbf{COMET}\\ 
\hline
Top-$1$ & $27.76$ & $76.38$\\
Top-$3$ & $38.55$ & $81.02$\\
Top-$5$ & $38.53$ & $81.01$\\
Top-$7$ & $\bf{38.90}$ & $\bf{81.14}$\\
\hline
    \end{tabular}
\caption{
    BLEU and COMET scores on the NTUML2021 CS validation set with $n$$=$$7$ and varying top-$k$ values.
	}\label{expert_performance_k}
\end{table}

\subsection{Experimental Setting}

\begin{table*}[]
	\renewcommand
	\arraystretch{1.2}
	\centering
\begin{tabular}{l|cccc}
\hline
\multirow{2}{*}{\textbf{Model}} & \multicolumn{2}{c}{\textbf{Fisher}} & \multicolumn{2}{c}{\textbf{NTUML2021}}\\
& \textbf{CS}       & \textbf{Mono.}     & \textbf{CS}       & \textbf{Mono.}    \\ \hline
Whisper                & $31.56$ / $72.63$      & $31.04$ / $75.93$      & $29.12$ / $74.90$    & $28.92$ / $77.54$ \\
SeamlessM4T            & $36.89$ / $76.13$    & $35.04$ / $79.71$     & $38.03$ / $79.93$    & $35.78$ / $81.17$     \\
Whisper+LLaMA            & $29.67$ / $70.99$    & $33.20$ / $76.08$     & $34.85$ / $79.08$ & $35.21$ / $\bf{81.68}$    \\
LLaST                  & $33.78$ / $75.35$      & $30.67$ / $78.37$      & $32.53$ / $78.92$    & $33.81$ / $79.03$ \\
Ours                   & $\bf{37.51}$ / $\bf{77.54}$    & $\bf{35.87}$ / $\bf{80.14}$     & $\bf{39.52}$ / $\bf{81.33}$    & $\bf{36.27}$ / $ 81.65$     \\ \hline
\end{tabular}
\caption{
    BLEU$\uparrow$ / COMET$\uparrow$ scores of various models on the Fisher and NTUML2021 test sets. CS refers to CS speech translation data, and Mono. refers to monolingual ST data. 
	}\label{tab_main_performance}
\end{table*}

\paragraph{Metrics}
We report SacreBLEU~\citep{post-2018-call} and COMET~\citep{DBLP:conf/emnlp/ReiSFL20} to evaluate translation quality. ASR performance is measured by Word Error Rate (WER)~\citep{morris2004and} for English and Spanish, and Character Error Rate (CER) for Chinese.

\paragraph{Datasets}

Following ~\citet{weller2022end} and ~\citet{yang2024investigating}, we select the Fisher and NTUML2021 datasets to evaluate model performance, as well as the widely used Common Voice dataset for ASR. The dataset details are as follows:

\begin{itemize}
\item{{\textbf {Common Voice.}}}~\citep{ardila-etal-2020-common} This is a widely used multilingual ASR dataset consisting of monolingual speech and corresponding transcriptions. We select the English, Spanish, and Chinese subsets as our ASR data.

\item{{\textbf {Fisher.}}}~\citep{cieri2004fisher} This is a CS speech translation dataset containing English–Spanish source speech with English target translations. We follow the splits defined by \citet{weller2022end} and adopt their separation of monolingual and CS speech translation data.

\item{{\textbf {NTUML2021.}}}~\citep{yang2024investigating} This dataset consists of Chinese–English source speech with English target translations. We adopt the original data splits and categorize samples whose transcriptions contain both Chinese and English as CS speech translation data, while the remaining samples are treated as monolingual speech translation data.

\end{itemize}

The numbers of parallel speech-text pairs and total hours for each dataset are reported in Table~\ref{dataset_statistics}.

\paragraph{Model Details.} 
We use Llama-3-8B-Instruct\footnote{\url{https://huggingface.co/meta-llama/Meta-Llama-3-8B-Instruct}} as the base LLM and adopt Whisper-large-v3 encoder as the speech encoder. The MoE speech projector consists of three layers and employs ReLU~\citep{agarap2018deep} as the activation function.

\paragraph{Hyperparameter Tuning.} 
We tune the number of experts per group $n$ and the top-$k$ value on the CS validation set of the NTUML2021 dataset. For consistency, we use a uniform number of experts across all groups to avoid introducing additional variability, although this number can be configured independently for each group.
Specifically, we first fix the top-$k$ value to $3$ and vary $n$ over the set $\{3, 5, 7, 9, 11\}$. As shown in Table ~\ref{expert_performance_n}, performance steadily improves as $n$ increases until it peaks at $n$$=$$7$, and gradually declines thereafter. We then fix $n$$=$$7$ and tune the top-$k$ value over the set $\{1, 3, 5, 7\}$. As illustrated in Table~\ref{expert_performance_k}, activating more experts generally improves performance, with top-$7$ achieving the best results.

Accordingly, we set $n$$=$$7$ and select the top-$7$ experts during inference.

\subsection{Baselines}
Our baselines include the following models:

\begin{itemize}

\item{{\textbf {Whisper.}}}~\citep{pmlr-v202-radford23a}. As a large-scale pretrained model, Whisper demonstrates strong performance across a wide range of speech translation tasks and is widely regarded as a robust baseline.

\item{{\textbf {SeamlessM4T.}}}~\citep{seamless2023}. Similar to Whisper, SeamlessM4T also exhibits strong performance across diverse speech translation tasks and serves as a competitive baseline.

\item{{\textbf {Whisper+LLaMA.}}} We construct a cascaded speech translation pipeline by combining Whisper for ASR with LLaMA for MT. Specifically, Whisper is first used to transcribe speech into source language text, which is then translated into the target language by LLaMA. This pipeline represents a strong cascaded baseline for comparison with our E2E speech translation models.

\item{{\textbf {LLaST.}}}~\citep{chen-etal-2024-llast}. As there is currently no work specifically targeting E2E CS speech translation with LLMs, we include LLaST as a representative baseline. It follows a mainstream architecture that integrates a pretrained speech encoder with an LLM and shows competitive performance on monolingual ST tasks.

\end{itemize}

\subsection{Main Results}
Table~\ref{tab_main_performance} presents the BLEU and COMET scores of various models on the Fisher and NTUML2021 test sets under two evaluation settings: CS speech translation data only (CS) and monolingual ST data only (Mono.).

Under the CS setting, our model achieves state-of-the-art performance with BLEU/COMET scores of $37.51/77.54$ on Fisher and $39.52/81.33$ on NTUML2021, clearly outperforming all baselines. Notably, although primarily designed for CS speech translation, our approach also delivers competitive or even superior performance in the Mono. setting, indicating its effectiveness for both monolingual and CS speech translation scenarios.

Overall, these results demonstrate the advantage of our method in capturing fine-grained speech features and improving translation robustness across diverse scenarios.
\begin{table}[!t]
	\renewcommand
	\arraystretch{1.2}
	\centering
\small
\begin{tabular}{l|cc}
\hline
\textbf{Model} & \textbf{BLEU} & \textbf{COMET}\\ 
\hline
Ours & $39.52$ & $81.33$  \\
\hdashline
\quad \emph{w/o MoE Speech Projector} & $36.85$ & $79.95$\\
\quad \emph{w/o $\mathcal{L}_{\text{lang}}$} & $39.05$ & $80.88$\\
\quad \emph{w/o $\mathcal{L}_{\text{balance}}$} & $38.86$ & $81.03$\\
\quad \emph{w/o $\mathcal{L}_{\text{transition}}$} & $39.21$ & $81.24$\\
\quad \emph{conventional $\mathcal{L}'_{\text{balance}}$} & $38.68$ & $80.79$\\
\quad \emph{w/o Stage 1} & $38.14$ & $80.31$\\
\quad \emph{w/o Stage 3} & $37.74$ & $80.02$\\

\hline
    \end{tabular}
\caption{
    Ablation studies on the NTUML2021 CS test set. 
	}\label{tab_ablation}
\end{table}
\subsection{Ablation Studies}
As shown in Table~\ref{tab_ablation}, we consider the following variants:
\begin{itemize}
\item{\emph{w/o MoE Speech Projector.}} In this variant, the MoE speech projector is replaced with a single MLP projector that adopts the same number of layers and ReLU activation. Accordingly, the language-specific loss and the intra-group load balancing loss are removed, along with Stage 2. As indicated in Line $2$, replacing the MoE speech projector with an MLP projector leads to a significant performance drop, highlighting the effectiveness of our proposed MoE speech projector in capturing fine-grained representations for CS speech.
\item{\emph{w/o $\mathcal{L}_{\text{lang}}$ or $\mathcal{L}_{\text{balance}}$.}} To validate the benefit of our carefully designed loss functions, we evaluate the variant in which $\mathcal{L}_{\text{lang}}$ or $\mathcal{L}_{\text{balance}}$ is excluded from Stages 2 and 3. As shown in Line $3$ and $4$, we find a notable performance degradation, indicating that these two auxiliary losses play an important role in guiding expert specialization as well as routing consistency.
\item{\emph{w/o $\mathcal{L}_{\text{transition}}$.}} We remove $\mathcal{L}_{\text{transition}}$ from Stages 3 and 4 to evaluate its impact. As illustrated in Line $5$, the resulting performance decline suggests that the transition loss facilitates smoother adaptation across different domains.
\item{\emph{conventional $\mathcal{L}'_{\text{balance}}$}.} For a direct comparison, we replace our proposed $\mathcal{L}_{\text{balance}}$ with the conventional $\mathcal{L}'_{\text{balance}}$ introduced by~\citet{fedus2022switch}:
\begin{equation}
\begin{aligned}
\mathcal{L}'_{\text{balance}}=\sum_{l=1}^L\sum_{i=1}^N{f'_{il} \cdot P'_{il}},
\end{aligned}
\end{equation} where $f'_{il} =\frac{1}{{T}} \sum_{T} \mathds{1}\{\arg\max_r\ p_r(h_l^t)=i\}$, and $P'_{il} = \frac{1}{{T}}\sum_{T} p_i(h_l^t)$.

This comparison allows us to assess how different balancing objectives affect expert utilization and training stability. Results in Line~$6$ show a slightly lower performance, suggesting that our customized balance loss better facilitates stable and effective expert utilization within our MoE framework.

\item{\emph{w/o Stage 1.}} For this variant, training starts from Stage 2 with all expert weights randomly initialized. As shown in Line~$7$, removing Stage 1 causes a noticeable decline in performance. In addition, we observe that the training loss decreases more slowly, suggesting that Stage~1 helps provide a better initialization for subsequent stages and contributes to improved final performance.

\item{\emph{w/o Stage 3.}} Under this variant, Stage~3 is omitted and Stage~4 is initialized using the weights from Stage~2. Moreover, the training pipeline is further simplified to a direct transition from monolingual ASR to CS speech translation data. As reported in Line~$8$, the notable performance degradation reveals a considerable task gap between ASR and CS speech translation, validating the necessity of Stage~3 for effective domain adaptation.

\end{itemize}

\begin{table}[!t]
	\renewcommand
	\arraystretch{1.2}
	\centering
\small
\begin{tabular}{l|llll}
\hline
\multicolumn{1}{c|}{\multirow{2}{*}{\textbf{Top-$k$}}} & \multicolumn{2}{c}{\textbf{Fisher}}                      & \multicolumn{2}{c}{\textbf{NTUML2021}}                          \\
\multicolumn{1}{c|}{}                       & \multicolumn{1}{c}{\textbf{En}} & \multicolumn{1}{c}{\textbf{Es}} & \multicolumn{1}{c}{\textbf{En}} & \multicolumn{1}{c}{\textbf{zh-CN}} \\ \hline
Top-1  & $97.62\%$ & $95.66\%$ & $99.99\%$ & $85.21\%$\\
\quad \emph{w/o $\mathcal{L}_{\text{lang}}$}   & $54.42\%$  & $51.16\%$ & $45.72\%$  & $69.28\%$ \\
\hdashline
Top-7  & $96.72\%$ & $95.76\%$ & $99.98\%$ & $81.03\%$\\
\quad \emph{w/o $\mathcal{L}_{\text{lang}}$}   & $51.06\%$  & $55.00\%$ & $45.66\%$  & $69.80\%$\\ \hline
\end{tabular}

\caption{
    Proportion of speech tokens routed to the corresponding language expert group within the top-$k$ experts.
	}\label{expert_routing}
\end{table}

\subsection{Analysis of Experts Assignment}
To further investigate how the MoE projector adaptively routes speech tokens to appropriate experts, we conduct an analysis using monolingual speech data from the Common Voice corpus, as oracle language labels are unavailable in CS speech. We use English and Spanish speech for the model trained on Fisher, whereas for the model trained on NTUML2021, English and Chinese speech are used. Specifically, for each speech token, we record the ranking of experts assigned by the router and compute the proportion of top-$k$ experts that belong to the corresponding language expert group. As shown in Table~\ref{expert_routing}, the majority of speech tokens are routed to their corresponding language expert groups, demonstrating that the MoE projector effectively captures cross-lingual semantic distinctions. Specifically, the routing accuracy exceeds $95\%$ for both English and Spanish tokens, while that for Chinese tokens is slightly lower but still above $80\%$, which is likely due to data imbalance across languages. Meanwhile, the proportion of correctly routed speech tokens for all languages drops significantly when $\mathcal{L}_{\text{lang}}$ is removed, indicating that $\mathcal{L}_{\text{lang}}$ plays a crucial role in guiding the router to learn appropriate expert assignments. In addition, we compute the expert ranks within the corresponding language expert group for each speech token, and report their mean and variance, which are $4.25$ / $0.69$ and $4.77$ / $0.88$ for the models trained on Fisher and NTUML2021, respectively. In contrast, these statistics become $4.14$ / $1.05$ and $4.67$ / $1.21$ when $\mathcal{L}_{\text{balance}}$ is excluded. These results indicate that $\mathcal{L}_{\text{balance}}$ substantially reduces the variance of correct expert ranks, thereby promoting more balanced routing.

\begin{figure}[t]
	\centering
	\includegraphics[width=\columnwidth]{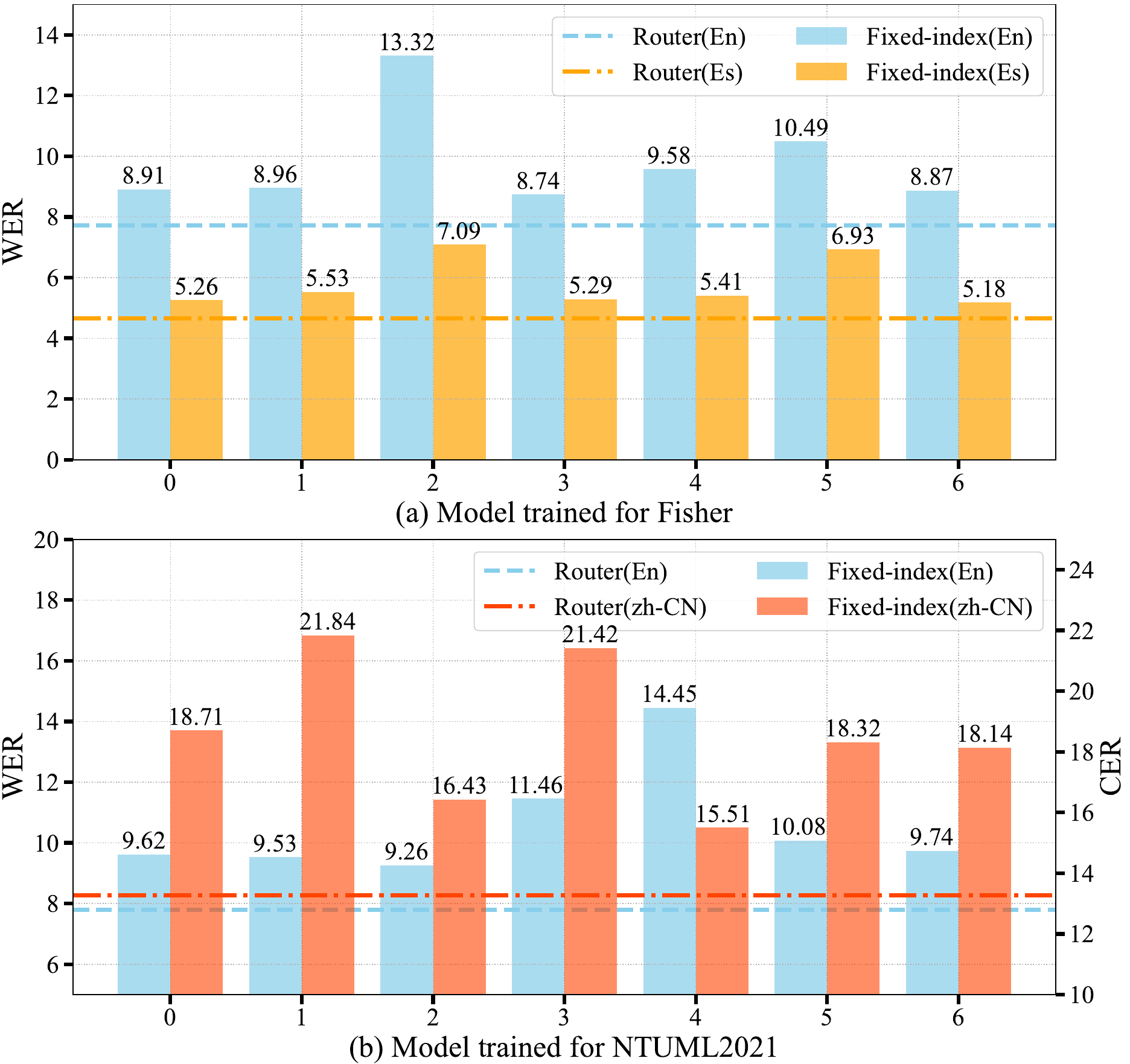}
 	\caption{Comparison of WER$\downarrow$(En, Es) and CER$\downarrow$(zh-CN) scores for the router and fixed-index variants. The x-axis represents the expert index. Dashed lines and bars indicate the performance of the router and fixed-index variants, respectively, and different colors denote different languages in the ASR test sets. }\label{combined_expert_performance}
\end{figure}

\subsection{Expert-Level Semantic Analysis}
In this set of experiments, we further analyze the semantic comprehension capability of individual experts using ASR data. Specifically, we disable the router and assign all speech tokens to a single expert, iterating over all expert indices within the corresponding language expert group. As presented in Figure~\ref{combined_expert_performance}, enabling the router consistently achieves the best performance across datasets. Meanwhile, the performance of fixed-index experts does not degrade substantially when operating independently. These results suggest that different experts capture complementary aspects of semantic understanding, and that our router can effectively and adaptively route speech tokens to appropriate experts.

\section{Conclusion and Future Work}
In this work, we first explore LLMs in end-to-end CS speech translation, and achieve significant improvements in two key aspects. First, we propose a novel MoE speech projector, which enables capturing fine-grained representations of CS speech. Second, we propose a novel multi-stage training paradigm that effectively utilizes diverse data domains while ensuring smooth transitions across training stages. Extensive experiments and analyses verify the effectiveness of our model.
For future work, we plan to extend our model to support a wider range of CS language pairs and to investigate more adaptive expert selection mechanisms. Besides, we aim to generalize our model to related CS tasks such as ASR and MT to further validate its robustness and generalizability.

\section*{Acknowledgments}
This work was supported by the First Batch of Projects for the 2025 ``Intergovernmental International Science, Technology and Innovation Cooperation'' of the National Key Research and Development Program of China under Grant 2025YFE0121700, Natural Science Foundation of Fujian Province of China (No. 2024J011001), and the Public Technology Service Platform Project of Xiamen (No.3502Z20231043). We also thank the reviewers for their insightful comments.

\bibliographystyle{named}
\bibliography{ijcai26}

\appendix

\begin{table*}[t]
	\renewcommand
	\arraystretch{1.3}
	\centering
\begin{tabular}{l|cccc}
\hline
\multirow{2}{*}{\textbf{Model}} & \multicolumn{2}{c}{\textbf{Fisher}} & \multicolumn{2}{c}{\textbf{NTUML2021}}\\
& \textbf{CS}       & \textbf{Mono.}     & \textbf{CS}       & \textbf{Mono.}    \\ \hline
SALMONN-13B                & $20.52$ / $72.21$ & $16.57$ / $68.95$ & $16.15$ / $71.25$ & $14.85$ / $71.53$ \\
Qwen2-Audio-7B-Instruct            & $21.23$ / $70.88$ & $21.49$ / $70.52$ & $23.66$ / $71.79$ & $27.22$ / $78.03$ \\
Qwen3-Omni-30B-A3B-Instruct            & $34.89$ / $79.24$ & $34.81$ / $79.70$ & $35.88$ / $80.09$ & $33.63$ / $\bf{82.42}$ \\
Voxtral-Mini-3B-2507            & $34.85$ / $79.19$ & $32.71$ / $78.17$ & $28.66$ / $77.50$ & $26.05$ / $77.38$ \\
Voxtral-Small-24B-2507           & $36.51$ / $\bf{79.86}$ & $34.65$ / $78.39$ & $32.20$ / $78.94$ & $27.71$ / $77.66$ \\
Ours                   & $\bf{37.51}$ / $77.54$    & $\bf{35.87}$ / $\bf{80.14}$     & $\bf{39.52}$ / $\bf{81.33}$    & $\bf{36.27}$ / $ 81.65$     \\ \hline
\end{tabular}
\caption{
    BLEU$\uparrow$ / COMET$\uparrow$ scores of various models on the Fisher and NTUML2021 test sets.
	}\label{tab_audiollm_performance}
\end{table*}

\begin{table*}[b]
	\renewcommand
	\arraystretch{1.2}
	\centering
\centering
\begin{CJK}{UTF8}{bsmi}
\begin{tabular}{cll}
\hline
\textbf{Case} & \textbf{Example 1} & \textbf{Example 2} \\ \hline

Transcription
& 好那這個 \textcolor{blue}{restricted boltzmann machine}
& 如果你真的有實做 \textcolor{blue}{vae} 或 \textcolor{blue}{flow} 的話 \\

Ground Truth
& Well then, this \textcolor{blue}{Restricted Boltzmann Machine}
& If you really have implemented \textcolor{blue}{VAE} or \textcolor{blue}{FLOW}  \\

Ours
& Okay, what about this \textcolor{blue}{Restricted} \textcolor{red}{Boltzman} \textcolor{blue}{Machine}?
& If you really want to implement \textcolor{blue}{VAE} or \textcolor{blue}{Flow} \\

SeamlessM4T
& Okay, this \textcolor{red}{restrictive positioning} \textcolor{blue}{machine}
& If you really have \textcolor{red}{10} \textcolor{red}{v1} or \textcolor{red}{v2} \\ \hline
\end{tabular}
\caption{
A case study of CS speech translation on the NTUML2021 test set. Words in \textcolor{blue}{blue} denote CS words in the source speech, while words in \textcolor{red}{red} indicate translation errors.
}\label{case_study_ntuml}
\end{CJK}
\end{table*}

\section{Implementation Details}
Our implementation is based on Xtuner\footnote{\url{https://github.com/InternLM/xtuner}}~\citep{2023xtuner}. We set the LoRA $\alpha$ value to $256$ for the LLM and to $64$ for the speech encoder. In Stages 1 and 2, our model is trained on ASR data from Common Voice in the corresponding languages. In Stage 3, we gradually transition the training data from ASR to monolingual ST data of the Fisher or NTUML2021 datasets. Finally, Stage 4 further transitions the model from monolingual ST to CS speech translation data of the Fisher or NTUML2021 datasets. Each stage is trained for a single epoch using the AdamW optimizer with a learning rate of $2\times10^{-4}$, and $\beta_1=0.9$, $\beta_2=0.999$. We set the per-device batch size to $8$ and apply gradient accumulation with $4$ steps, resulting in an effective batch size of $32$. During inference, we employ beam search with a beam size of $5$. All experiments are conducted on NVIDIA A100 GPUs, and a full training cycle takes about 2 days on 4 GPUs. We keep all training and inference hyperparameters identical to those of the baselines except for the training strategy. For end-to-end baselines, we directly fine-tune the models on the combined monolingual and CS speech translation data. For cascade baselines, the ST data are split into ASR and MT subsets, which are used to fine-tune the ASR and MT models independently. 

\section{Comparison with Audio-LLMs}
We further compare our model with recent Audio-LLMs, including SALMONN~\citep{tang2024salmonn}, Qwen2-Audio~\citep{Qwen2-Audio}, Qwen3-Omni~\citep{Qwen3-omni}, and Voxtral~\citep{liu2025voxtral}, in a zero-shot setting. As shown in Table~\ref{tab_audiollm_performance}, our model consistently achieves the best BLEU scores across all settings and remains competitive in terms of COMET, only slightly underperforming in two cases compared to significantly larger models. Overall, our model still achieves superior or highly competitive performance against recent Audio-LLMs.

\section{Case Study}
To better analyze the behavior of the model in CS scenarios, we present two representative examples from our model and the best baseline, SeamlessM4T, on the NTUML2021 test set. As shown in Table~\ref{case_study_ntuml}, example 1 contains an utterance with CS words \textit{restricted boltzmann machine}. Our model generates a nearly correct translation with only a minor spelling mistake \textit{Boltzman}, where the letter \textit{n} is missing. In contrast, SeamlessM4T translates them as \textit{restrictive positioning machine}, which is semantically irrelevant but phonetically similar to the source speech. In example 2, the speech includes the CS words \textit{vae} and \textit{flow}. Our model successfully preserves these words in the translation. However, SeamlessM4T outputs \textit{v1} and \textit{v2}. Meanwhile, the error word \textit{10} originates from the Chinese character \begin{CJK}{UTF8}{bsmi}\textit{實}\end{CJK}, whose pronunciation is similar to \textit{ten} in Chinese. These two examples reveal that our model exhibits stronger capability in modeling CS speech and aligning speech with appropriate textual representations, while SeamlessM4T tends to rely on acoustic similarity and fails to adequately separate features across languages, leading to semantically inconsistent translations of CS words.

\section{Computational Overhead and Efficiency.}

To evaluate the computational overhead and inference efficiency of the proposed MoE speech projector, we first report the number of model parameters. Compared with the model without MoE speech projector, the total number of parameters increases from $9.42$B to $9.93$B for the model with MoE speech projector, corresponding to an approximate $5\%$ increase.

We further evaluate inference efficiency on the NTUML test set. The model with MoE speech projector achieves Real-Time Factor (RTF$\downarrow$) and tokens per second (Tok/Sec$\uparrow$) values of $0.68$ / $5.72$, compared with $0.69$ / $5.96$ for the model without MoE speech projector. Although the MoE module introduces additional parameters, the RTF is slightly improved because the model without MoE speech projector tends to generate longer outputs, leading to increased decoding time. In terms of decoding throughput, the MoE variant introduces only a minor slowdown of approximately $4\%$.

Overall, the proposed MoE speech projector introduces relatively limited computational and memory overhead while maintaining comparable inference efficiency.

\begin{table}[h]
	\renewcommand
	\arraystretch{1.2}
	\centering
\small
\begin{tabular}{ccc}
\hline
\textbf{Number of CS Words} & \textbf{BLEU} & \textbf{COMET} \\ \hline
1                           & 39.89         & 82.04          \\
2                           & 39.03         & 81.02          \\
3                           & 39.68         & 80.82          \\
4+                          & 40.56         & 81.31          \\ \hline
\end{tabular}
\caption{
    Translation performance on the NTUML2021 CS test set with respect to the number of CS words in each utterance.
	}\label{performance_cs_words}
\end{table}

\section{Robustness to CS Word Quantity}
To further investigate the robustness of our model in CS scenarios, we conduct an analysis experiment that examines how translation performance varies with the number of CS words in each utterance. Specifically, the NTUML2021 dataset is adopted due to its clear boundary between Chinese and English words, and test samples are grouped according to the number of CS words contained in each utterance. Considering that sentences with four or more CS words are relatively scarce (approximately $100$ samples in total), we merge them into a single group to ensure reliable performance estimates. Despite the increasing linguistic complexity introduced by more CS words, the results in Table~\ref{performance_cs_words} show that our model maintains stable performance without significant degradation, demonstrating the effectiveness and robustness of our approach in challenging CS scenarios.

\end{document}